\documentclass[letterpaper, 10 pt, journal, twoside]{IEEEtran}
\usepackage{amsmath,amsfonts}
\usepackage{algorithmic}
\usepackage{algorithm}
\usepackage{array}
\usepackage[caption=false,font=normalsize,labelfont=sf,textfont=sf]{subfig}
\usepackage{stfloats}
\usepackage{multirow}
\usepackage{rotating}
\usepackage{url}
\usepackage{verbatim}
\usepackage{graphicx}
\usepackage{cite}
\usepackage{hyperref}
\hypersetup{hidelinks,
	colorlinks=true,
	urlcolor=magenta,
	pdfstartview=Fit,
	breaklinks=true}
\begin{document}

\title{Towards Better Data Exploitation in Self-Supervised Monocular Depth Estimation}

\author{Jinfeng Liu, Lingtong Kong, Jie Yang,  and Wei Liu
\thanks{Manuscript received: August, 6, 2023; Revised: October, 18, 2023; Accepted: November, 20, 2023. This letter was recommended for publication by Associate Editor I. Gilitschenski and Editor C. Cadena Lerma upon evaluation of the reviewer's comments. This work was supported by NSFC under Grant 62376153, and in part by Pujiang Program under Grant 22PJ1406600. (\textit{Corresponding authors: Jie Yang and Wei Liu.})}
\thanks{The authors are with the Institute of Image Processing and Pattern Recognition, Department of Automation, Shanghai Jiao Tong University, Shanghai, China (e-mail:
ljf19991226@sjtu.edu.cn;	ltkong218@gmail.com;
jieyang@sjtu.edu.cn; weiliucv@sjtu.edu.cn).}
\thanks{Digital Object Identifier 10.1109/LRA.2023.3337594}
}

\markboth{IEEE Robotics and Automation Letters. Preprint Version. Accepted November, 2023}
{Liu \MakeLowercase{\textit{et al.}}: Towards Better Data Exploitation in Self-Supervised Monocular Depth Estimation}



\maketitle

\begin{abstract}
Depth estimation plays an important role in robotic perception systems. The self-supervised monocular paradigm has gained significant attention since it can free training from the reliance on depth annotations. Despite recent advancements, existing self-supervised methods still underutilize the available training data, limiting their generalization ability. In this paper, we take two data augmentation techniques, namely \textit{Resizing-Cropping} and \textit{Splitting-Permuting}, to fully exploit the potential of training datasets. Specifically, the original image and the generated two augmented images are fed into the training pipeline simultaneously and we leverage them to conduct self-distillation.  Additionally, we introduce the detail-enhanced DepthNet with an extra full-scale branch in the encoder and a grid decoder to enhance the restoration of fine details in depth maps. Experimental results demonstrate our method can achieve state-of-the-art performance on the KITTI and Cityscapes datasets. Moreover, our KITTI models also show superior generalization performance when transferring to Make3D, NYUv2 and Cityscapes datasets. Our codes are available at \url{https://github.com/LiuJF1226/BDEdepth}.
\end{abstract}


%
\IEEEpeerreviewmaketitle

\section{Introduction}
\IEEEPARstart{D}{epth} estimation is one of the critical tasks in robotics, which can help robots perceive the scene structure and navigate in the 3D world. Due to the simplicity, flexibility and low cost of monocular cameras, monocular depth estimation recently has become a research hotspot in the realm of robotics and computer vision. Nonetheless, amassing extensive annotated training datasets for supervised learning \cite{eigen,liu,li,laina,fu,eigen2,etc,dsl} is extremely expensive. To reduce the reliance on labeled data, self-supervised paradigm \cite{Garg, Godard, sfmlearner} has emerged as a viable approach, leveraging monocular videos, stereo pairs, or even both to provide supervisory signals via novel view synthesis. This paradigm frames training as minimizing photometric reprojection error, which is also the core idea of unsupervised flow \cite{MDFlow} and direct visual odometry (DVO) in simultaneous localization and mapping (SLAM) systems.

Recent self-supervised monocular depth estimation methods \cite{geonet, sdo, poggi, monodepth2, packnet,hrdepth,r-msfm,adaadepth, brnet,Cha,lite-mono,planedepth,devnet,diffnet,radepth,scdepth,epcdepth,Pilzer,Bello,direct, superdepth, jiang2020, li2021,Wagstaff,manydepth} have achieved remarkable results, but they still fail to make full use of the potential information within the training datasets. Consequently the generalization ability is limited, which is significant in robotics since robots should adapt to diverse scenarios. Data augmentation is a common and cost-effective way in computer vision to improve the generalization ability of the model. However, most existing methods just simply take horizontal flipping and color jittering. Although some \cite{planedepth,scdepth} also use resizing and cropping, they only train with augmented images and discard the original images. In order to fully dig into the potential of training datasets for various robotic applications, we use two data augmentation approaches, aforementioned \textit{Resizing-Cropping} and novel \textit{Splitting-Permuting}. The latter first splits an image into patches and then permutes them to generate a new image with context broken. Hence, there are two augmented images generated from each target image. Each of the three image views participates in the view synthesis of the self-supervised training process. And depth maps from the augmented views are also used for self-distillation with the original one.

On the other hand, most approaches follow Monodepth2 \cite{monodepth2} to adopt a prevalent encoder-decoder architecture for depth estimation. The encoder extracts five-level features from the input image. The features encapsulate both global and local information of the image, with varying receptive fields at different scales, which are subsequently fed into a U-Net \cite{unet} decoder to generate the depth map. This architecture empowers Monodepth2 to gain advanced performance. However, it still struggles to recover fine details in depth maps. If the depth map is rough, robots could miss details or misinterpret object relationships when mapping, leading to inaccurate environmental perception. To tackle this limitation, we propose a detail-enhanced DepthNet in this paper, with an extra full-scale branch in the encoder and a grid decoder.

Our main contributions can be summarized as follows:
\begin{itemize}
\item{To fully excavate the potential of datasets for better generalization performance, we use classic \textit{Resizing-Cropping} and  novel \textit{Splitting-Permuting} augmentation to generate another two augmented views for self-distillation.}
\item{To enhance the fine details of depth maps, we propose to add a full-scale branch and adopt a grid decoder, forming our detail-enhanced DepthNet.
}
\item{Experiments show that we achieve state-of-the-art performance on the KITTI \cite{kitti} and Cityscapes \cite{city} datasets. And our KITTI models also have better performance when generalizing to Make3D \cite{make3d}, NYUv2 \cite{nyu} and Cityscapes datasets.}
\end{itemize}

\section{Related Work}
\IEEEpubidadjcol
\subsection{Supervised Monocular Depth Estimation}
Fully supervised methods for depth estimation  in the field of robotics or autonomous driving require depth ground truth when training. Among this category of methods, Eigen et al. \cite{eigen} first propose to use convolutional neural network (CNN) for monocular depth estimation. They employ a multi-scale neural network to estimate the depth map. Liu et al. \cite{liu} combine a depth CNN with continuous conditional random fields (CRFs) for depth estimation from a single image. Li et al. \cite{li} propose a multi-scale approach, which uses a deep neural network to regress depth at the super-pixel scale and applies CRFs for post-processing. Laina et al. \cite{laina} introduce an innovative up-sampling module and utilize the reverse Huber loss to improve training. And Fu et al. \cite{fu} employ a multi-scale network and regard depth estimation as an ordinal regression task, resulting in improved accuracy and faster convergence.

\subsection{Self-supervised Monocular Depth Estimation}
The high expense of collecting large annotated datasets for depth estimation has led to the emergence of self-supervised methods that do not rely on the ground truth. Garg et al. \cite{Garg} first consider the training of monocular depth estimation as a view synthesis problem  with a photometric-consistency loss between stereo pairs. Building upon this idea, Godard et al. \cite{Godard} introduce a left-right disparity consistency loss, which can improve the accuracy of depth estimation. For training with monocular video frames, Zhou et al. \cite{sfmlearner} propose a pioneering method which jointly learns monocular depth and ego-motion from monocular videos in a self-supervised way. They also use an explainability prediction network to exclude pixels that violate view synthesis assumptions, to improve the robustness when meeting with occlusion and moving objects. For dynamic scenes, other works introduce multi-task learning such as optical flow estimation \cite{geonet} and semantic segmentation \cite{sdo}, or introduce additional constraints, such as uncertainty estimation \cite{poggi}. Then Godard et al. \cite{monodepth2} propose Monodepth2, which employs a minimum reprojection loss to mitigate occlusion issues and an auto-masking loss to filter out moving objects with the similar velocity as the camera. This work achieves competitive results without introducing extra tasks, which presents promising prospects for robotic perception. Therefore many subsequent methods including ours follow and base on it. 

\subsection{Data Augmentation and Self-distillation}
For data augmentation, many existing methods just simply utilize horizontal flipping and color jittering. Some approaches \cite{scdepth,planedepth,radepth} additionally use resizing and cropping. And Peng et al. \cite{epcdepth} proposes a data augmentation approach called data grafting, which forces the model to explore more cues to infer depth besides the vertical image position. Among them, \cite{planedepth}, \cite{scdepth} and \cite{epcdepth} only utilize augmented images for training. Especially, PlaneDepth \cite{planedepth} achieves striking performance when using stereo pairs, but poor with monocular videos. The most similar to our work is RA-Depth \cite{radepth}, which uses resizing and cropping to generate images with arbitrary scales for the same scene. It modifies the camera intrinsics to keep the world coordinates, whereas we follow the assumption in \cite{planedepth} that all input images, including augmented views and original one, are captured by the same camera system.

Self-distillation is commonly used in self-supervised learning which means using a pretrained teacher network to generate pseudo labels for training a better student network. In depth estimation, for example, Pilzer et al. \cite{Pilzer} obtain multiple predictions from teacher network to generate pseudo labels for the student network training. Self-distillation can also be used to reduce the uncertainty of depth prediction \cite{Bello}, and solve the artifact problem caused by occlusion especially under stereo training \cite{planedepth}, which actually feeds a flipped version of the original image into a depth network to synthesize pseudo labels. In our work, we also use two data augmentation methods and generate pseudo labels from augmented images.

\section{Methodology}
\subsection{Preliminaries}
Self-supervised monocular depth estimation paradigm aims at training a depth network $\theta_{\rm{depth}}$ to predict a per-pixel depth map $D$ from the input RGB image $I$, without providing any depth ground truth. Specifically, considering a target view $I_t$ in a consecutive image sequence from a monocular video, the neighbour frames are used as the source views $\{I_s\}_{s=1}^{N}$ to carry out view synthesis. For each source view $I_s$, an auxiliary pose network $\theta_{\rm{pose}}$ is utilized to estimate the relative camera pose $T_{t \rightarrow s}=[R_{t \rightarrow s}|t_{t \rightarrow s}]$, where $R_{t \rightarrow s}$ and $t_{t \rightarrow s}$ represent the rotation and translation component respectively. Note that the pose network is only used for training and could be discarded in the inference, and it is not needed when training with only stereo pairs since the stereo baseline is already known. Then the target depth map $D_t$ and the relative camera pose $T_{t \rightarrow s}$ from the target view to the source view can be derived from:
\begin{equation}
D_t=\theta_{\rm{depth}}(I_t), ~~~ T_{t \rightarrow s}=\theta_{\rm{pose}}(I_t,I_s).
\end{equation}
And the scene at the viewpoint of $I_t$ can be synthesized from the source view $I_s$ as:
\begin{equation}
I_{s \rightarrow t}=I_s\big\langle proj(D_t,T_{t \rightarrow s},K) \big\rangle,
\end{equation}
where $I_{s \rightarrow t}$ is the synthesized image, $\langle \rangle$ is the sampling operator, $proj()$ returns the 2D coordinates of the depths in $D_t$ when reprojected into the
viewpoint of $I_s$, and $K$ is the camera intrinsics matrix assumed to be known. Finally the per-pixel photometric loss between $I_{s \rightarrow t}$ and $I_t$ is used to optimize the full network. We follow \cite{monodepth2} to use the combinaton of structural similarity (SSIM) \cite{ssim} and L1 distance as the photometric error function and take the minimum across all the source views at each pixel, which is formulated as:
\begin{equation}
L_{pe}=\min\limits_{s} pe(I_t,I_{s \rightarrow t}),
\end{equation}
\begin{equation}
pe(I_a,I_b)=\frac{\alpha}{2}(1-{\rm{SSIM}}(I_a,I_b)) + (1-\alpha)||I_a-I_b||_1,
\end{equation}
where $\alpha=0.85$. And also, the edge-aware smoothness loss is used to cope with depth discontinuities:
\begin{equation} \label{eq5}
L_{sm}=|\partial_x d_t^*|e^{-|\partial_x I_t|}+|\partial_y d_t^*|e^{-|\partial_y I_t|},
\end{equation}
where $d_t^*= d_t/\overline{d_t}$ is the mean-normalized inverse depth from \cite{direct} to discourage shrinking of the estimated depth.

\begin{figure*}[!t]
\centering
\includegraphics[width=7in]{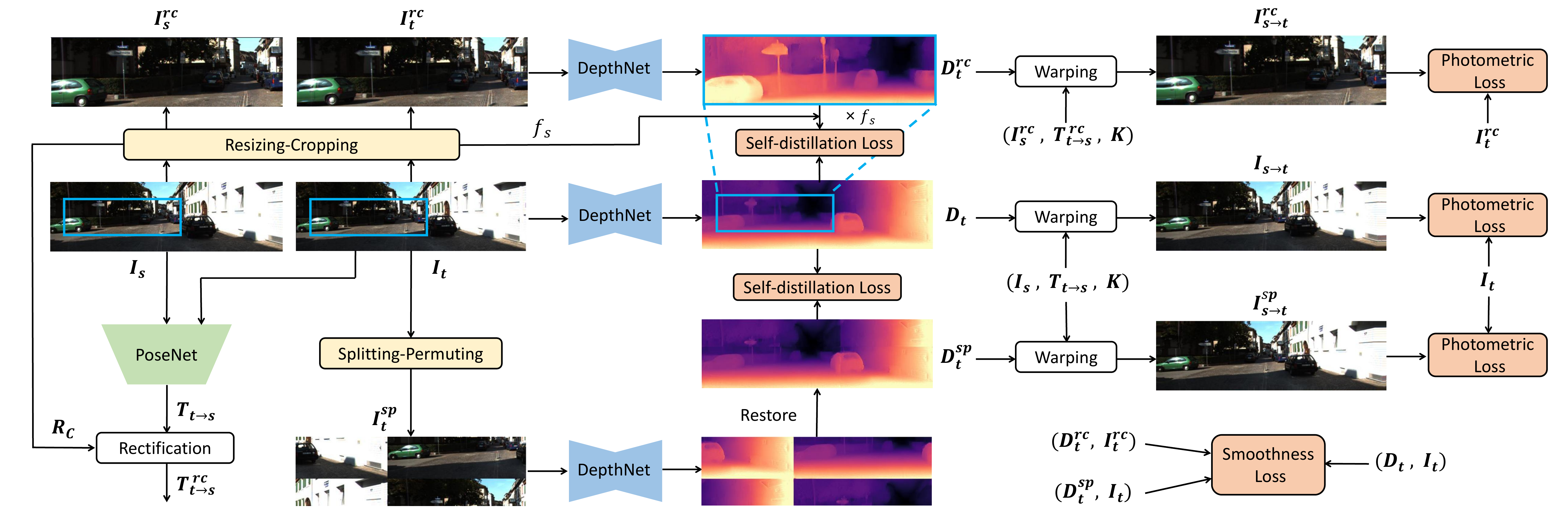}
\caption{The overview of our training pipeline. Two data augmentation strategies, \textit{Resizing-Cropping} and  \textit{Splitting-Permuting}, are used to generate another two augmented views $I_t^{sp}$ and $I_t^{rc}$ from the original target image $I_t$. Both the original view and the augmented views are involved in the standard self-supervised training. Additionally, we utilize the depth maps from the two augmented images as pseudo labels to provide more supervisory signals.}
\label{overview}
\end{figure*}

\subsection{Data Augmentation for Self-distillation}
In this paper, We adopt two data augmentation strategies, \textit{Resizing-Cropping} and  \textit{Splitting-Permuting}, to generate another two augmented views $I_t^{rc}$ and $I_t^{sp}$ from the original target image $I_t$ for self-distillation. It is worth noting that both the original image and augmented images are involved in the training process. The augmentation detail for a single image view is provided in Algorithm~\ref{alg1} and the overview of our training pipeline is depicted in Fig.~\ref{overview}.

\begin{algorithm}[H]
\caption{Data Augmentation}
\begin{algorithmic}
\STATE \textbf{Input}: An image $I$ of shape $(c, h, w)$, which is already resized to a specific input resolution $h \times w$.
\STATE \textbf{Output}: $I^{rc}$ through \textit{Resizing-Cropping} augmentation and $I^{sp}$ through \textit{Splitting-Permuting} augmentation.

\STATE \textbf{Generating $I^{rc}$}:
\STATE \hspace{0.5cm}Random sampling  scale factor $f_s$ from  [1.2, 2.0];
\STATE \hspace{0.5cm}$T \leftarrow Resize(I)$ to $(c, h\times f_s, w\times f_s)$;
\STATE \hspace{0.5cm}$I^{rc} \leftarrow RandomCrop(T)$ to $(c, h, w)$;

\STATE \textbf{Generating $I^{sp}$}:
\STATE \hspace{0.5cm}Random sampling the horizontal and vertical split ratios $r_h$ and $r_v$ from the range [0.1, 0.9];
\STATE \hspace{0.5cm}$T=I,~I^{sp}=I$;
\STATE \hspace{0.5cm}$T[:,h-int(h\times r_h):,:]\leftarrow I[:,:int(h\times r_h),:]$;
\STATE \hspace{0.5cm}$T[:,:h-int(h\times r_h),:]\leftarrow I[:,int(h\times r_h):,:]$;
\STATE \hspace{0.5cm}$I^{sp}[:,:,w-int(w\times r_v):]\leftarrow T[:,:,:int(w\times r_v)]$;
\STATE \hspace{0.5cm}$I^{sp}[:,:,:w-int(w\times r_v)]\leftarrow T[:,:,int(w\times r_v):]$;

\STATE \textbf{Return} $I^{rc}$, $I^{sp}$.
\end{algorithmic}
\label{alg1}
\end{algorithm}
\vspace{-2mm}

\subsubsection{Resizing-Cropping} We follow \cite{planedepth} to leverage an essential monocular cue that the closer an object is, the larger its relative size is. Therefore, it is assumed that when a image is scaled by a factor $f_s>1$, the relative size of an object also increases by $f_s$ and its depth decreases by $f_s$. This augmentation is applied to both target image and source images, generating $I_t^{rc}$ from $I_t$ and $I_s^{rc}$ from $I_s$. As proposed in \cite{planedepth}, given the original relative pose $T_{t \rightarrow s}=[R_{t \rightarrow s}|t_{t \rightarrow s}]$, the pose $T_{t \rightarrow s}^{rc}=[R_{t \rightarrow s}^{rc}|t_{t \rightarrow s}^{rc}]$ from $I_t^{rc}$ to $I_s^{rc}$ should be rectified as:
\begin{equation}
R_{t \rightarrow s}^{rc}=R_C R_{t \rightarrow s} R_C^{-1}, ~~~t_{t \rightarrow s}^{rc}=R_C t_{t \rightarrow s},
\end{equation}
where $R_C$ is the rectification matrix which transforms the original world coordinates to the new coordinates originated at the augmented image. 
Denote the center coordinate in the original image coordinate system as $(c_x, c_y)$, the center of cropped patch as $(p_x, p_y)$, the camera focal length as $(f_x, f_y)$, then $R_C$ can be written as:
\begin{equation}
R_C=\begin{bmatrix} 1 & 0 & (c_x-p_x)/f_x\\
0 & 1 & (c_y-p_y)/f_y \\
0 & 0 &  f_s
\end{bmatrix}.
\end{equation}
And the corresponding synthesized image $I_{s \rightarrow t}^{rc}$ from $I_{s}^{rc}$ and the photometric loss $L_p^{rc}$ can be obtained by:
\begin{equation}
I_{s \rightarrow t}^{rc}=I_s^{rc}\big\langle proj(D_t^{rc},T_{t \rightarrow s}^{rc},K)  \big\rangle,  
\end{equation}
\begin{equation}
L_{pe}^{rc}=\min\limits_{s} pe(I_t^{rc},I_{s \rightarrow t}^{rc}),
\end{equation}
where $D_t^{rc}=\theta_{depth}(I_t^{rc})$ is the predicted depth map from $I_t^{rc}$.

\subsubsection{Splitting-Permuting} Monocular depth estimation can easily get stuck in overfitting since its excessive dependence on the vertical image position \cite{epcdepth}. To improve the generalization ability, we propose a novel augmentation approach, \textit{Splitting-Permuting}, which provides perturbation and breaks the image context both vertically and horizontally. Specifically, the original image is firstly split into top part and bottom part with a random split ratio and the two parts are permuted to generate a new image. Then it is successively split into left and right parts and permuted, finally generating another augmented view. This is only conducted on the target image $I_t$ to produce $I_t^{sp}$, from which we can obtain the depth map as $D_t^{sp}=Restore(\theta_{depth}(I_t^{sp}))$. The operation $Restore()$ can restore the predicted map to the original context structure for reprojection, as shown in Fig.~\ref{overview}. And we use the original $I_t, I_s$ and the restored ${D}_t^{sp}$ to do view synthesis and calculate photometric loss as:
\begin{equation}
I_{s \rightarrow t}^{sp}=I_s\big\langle proj({D}_t^{sp},T_{t \rightarrow s},K)  \big\rangle,  
\end{equation}
\begin{equation}
L_{pe}^{sp}=\min\limits_{s} pe(I_t,I_{s \rightarrow t}^{sp}).
\end{equation}

\subsubsection{Self-distillation} To fully leverage the valid information from data augmentation, the depth maps predicted from augmented views are used to conduct self-distillation, which means we utilize ${D}_t^{rc}$ and ${D}_t^{sp}$ from \textit{Resizing-Cropping} and \textit{Splitting-Permuting} respectively as pseudo labels to provide more supervisory signals. Specifically, for ${D}_t^{rc}$, we first select the corresponding area $\widetilde{D}_t$ in $D_t$, from which the ${I}_t^{rc}$ is cropped. And then ${D}_t^{rc}$ is resized to the same resolution as $\widetilde{D}_t$, denoted as $\widetilde{D}_t^{rc}$. Therefore, this part of self-distillation loss is formulated as:
\begin{equation}
L_{sd}^{rc}={\rm SI}(\widetilde{D}_t, f_s \widetilde{D}_t^{rc}),
\end{equation}
where $\rm SI()$ is the widely used scale-invariant error function proposed in \cite{eigen}. Here the scale factor $f_s$ is due to aforementioned assumption that depth from original image is $f_s$ times of that from resized image. Then let $E=log(\widetilde{D}_t)-log(f_s \widetilde{D}_t^{rc})$ and $e_i$ denote the $i_{th}$ pixel of $E$, the scale-invariant loss can be calculated as:
\begin{equation}
{\rm SI}(\widetilde{D}_t, f_s \widetilde{D}_t^{rc})=\frac{1}{N} \sum_i e_i^2 - \frac{\beta}{N^2}\left(\sum_i e_i\right)^2,
\end{equation}
where $\beta=0.5$ and $N$ is the number of pixels. And for ${D}_t^{sp}$, we can similarly obtain the self-distillation loss as:
\begin{equation}
L_{sd}^{sp}={\rm SI}(D_t, {D}_t^{sp}).
\end{equation}

\begin{figure*}[!t]
\centering
\includegraphics[width=6.2in]{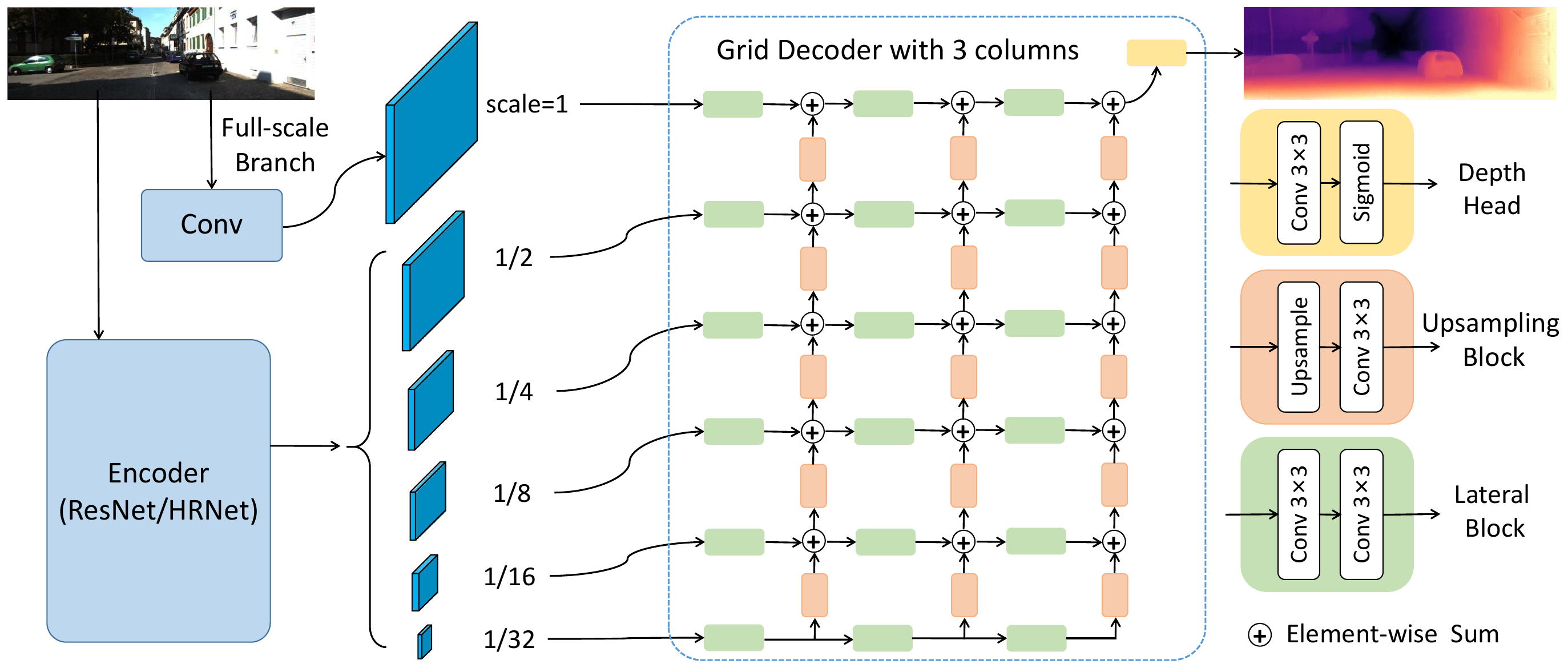}
\caption{The architecture of our proposed detail-enhanced DepthNet. There is a full-scale branch separated from the original encoder. So a six-level feature pyramid is generated and fed to the decoder. And we use a grid decoder instead of U-Net.}
\vspace{-2mm}
\label{arch}
\end{figure*}

\subsubsection{Loss Function} 
The standard self-supervised loss for the original image contains the photometric loss and the edge-aware smoothness loss:
\begin{equation}
L_{ss}=\mu L_{pe} + \gamma L_{sm},
\end{equation}
where $\gamma=0.001$ and $\mu$ is the auto-masking introduced in \cite{monodepth2}. This loss is applied to both two augmented views, formuating $L_{ss}^{rc}$ and $L_{ss}^{sp}$ as:
\begin{equation}
L_{ss}^{rc}=\mu L_{pe}^{rc} + \gamma L_{sm}^{rc},
\end{equation}
\begin{equation}
L_{ss}^{sp}=\mu L_{pe}^{sp} + \gamma L_{sm}^{sp},
\end{equation}
where $L_{sm}^{rc}$ and $L_{sm}^{sp}$ are the corresponding smoothness loss terms for the augmented views and can be calculated similarly as (\ref{eq5}). Considering both self-supervised losses and self-distillation losses, we can write the total loss function as:
\begin{equation}
L = \frac{1}{3}(L_{ss}+L_{ss}^{rc}+L_{ss}^{sp})+\lambda (L_{sd}^{rc}+ L_{sd}^{sp}),
\end{equation}
where $\lambda$ is set to 0.07 in our experiments.

\subsection{Detail-enhanced DepthNet}  
To enhance the fine details in depth maps, in this paper we propose to add a full-scale branch to the encoder and use grid-style decoder, forming the detail-enhanced DepthNet, as illustated in Fig.~\ref{arch}.

\subsubsection{Full-scale Branch} 
One reason can cause the loss of details is that the encoder preserves features with a maximum scale of $\frac{1}{2}$ and the decoder upsamples from $\frac{1}{2}$ scale to get the depth map at the full scale. BRNet \cite{brnet} obtains detailed information by reducing the stride of the first convolution, to maintain the full scale at the first feature level. Nevertheless, this will double calculation amount. For less computational overhead, we introduce a full-scale  branch separated from the original encoder, which is actually a convolution layer copied from the stem convolution of the encoder backbone, with the stride adjusted to 1. Hence, we can additionally preserve a feature at the full scale, constructing a six-level feature pyramid, which can provide more detailed information for the decoder to restore the depth map.

\subsubsection{Grid Decoder} Another point lies in that we replace the U-Net \cite{unet} decoder with a grid-style decoder, inspired from GridNet \cite{gridnet}. As shown in Fig.~\ref{arch} , each row consists several lateral blocks to form a stream, within which the feature resolution keeps constant. Each row operates at a different scale, and the columns of the decoder connect these row streams, allowing information transfer through upsampling blocks. Compared to U-Net architecture, our grid decoder can autonomously learn how to integrate information at different scales and use low-resolution information to guide high-resolution predictions, which is  beneficial to the detail restoration in pixel-wise tasks like depth estimation. And the columns (or lateral blocks at each row) of the grid decoder is set to 3 in this paper.

\begin{table*}
\begin{center}
\caption{Comparison results ( $640\times192$) on KITTI Benchmark with Raw ground truth}
\label{tab1}
\setlength{\tabcolsep}{1.5mm}{
\begin{tabular}{|c|c|c|c||c|c|c|c|c|c|c|}
\hline
 Method & Backbone & \#Params & Type & Abs Rel $\downarrow$ & Sq Rel $\downarrow$ & RMSE $\downarrow$ &RMSE log $\downarrow$ & $\delta<1.25\uparrow$ & $\delta<1.25^2\uparrow$ & $\delta<1.25^3\uparrow$\\
\hline
\hline
 Monodepth2~\cite{monodepth2}& ResNet18 & 14.3M & M & 0.115 & 0.903 & 4.863 & 0.193 & 0.877 & 0.959 & 0.981\\
 Monodepth2~\cite{monodepth2} & ResNet50& 32.5M & M & 0.110 & 0.831 & 4.642 & 0.187 & 0.883 & 0.962 & 0.982\\
 PackNet-SfM~\cite{packnet} & PackNet& 128M & M & 0.111 & 0.785 & 4.601 & 0.189 & 0.878 & 0.960 &  0.982 \\
 HR-Depth~\cite{hrdepth} & ResNet18& 14.6M & M & 0.109 & 0.792 & 4.632 & 0.185 & 0.884 & 0.962 & 0.983 \\
 R-MSFM6~\cite{r-msfm} & ResNet18& 3.8M & M & 0.112 & 0.806 & 4.704 & 0.191 & 0.878 & 0.960 & 0.981 \\
 ADAADepth~\cite{adaadepth} & ResNet18 & -& M & 0.111 & 0.817 & 4.685 & 0.188 & 0.883 & 0.961 & 0.982 \\
 BRNet~\cite{brnet} & ResNet18 & 19M & M & 0.105 & 0.698 & 4.462 & 0.179 & 0.890 & 0.965 & 0.984 \\
 Cha et al.~\cite{Cha} & ResNet18 & - & M & 0.110 & 0.763 & 4.648 & 0.184 & 0.877 & 0.961 & 0.983 \\
 Lite-Mono~\cite{lite-mono} & Lite-Mono & \textbf{3.1M} & M & 0.107 & 0.765 & 4.561 & 0.183 & 0.886 & 0.963 & 0.983 \\
 PlaneDepth~\cite{planedepth} & ResNet50 & - & M & 0.113 & 1.049 & 4.943 & - & 0.859 & - & -\\
 DevNet~\cite{devnet} & ResNet50 & - & M & 0.100 & 0.699 & 4.412 & 0.174 & 0.893 & 0.966 & \textbf{0.985}\\
 DIFFNet~\cite{diffnet} & HRNet18 & 10.9M & M & 0.102 & 0.764 & 4.483 & 0.180 & 0.896 & 0.965 & 0.983 \\
 RA-Depth~\cite{radepth} & HRNet18 & 10M & M & 0.096 & 0.632 & 4.216 & 0.171 & 0.903 & \textbf{0.968} & \textbf{0.985} \\
 \textbf{Ours} & ResNet18 & 18M& M & 0.102  & 0.697 & 4.454 & 0.178 & 0.890  & 0.965 & 0.984\\
 \textbf{Ours} & HRNet18 & 10.2M & M & \textbf{0.095}  & \textbf{0.621} & \textbf{4.183} & \textbf{0.170} & \textbf{0.904}  & \textbf{0.968} & \textbf{0.985}\\
\hline 
 Monodepth2~\cite{monodepth2} & ResNet18 & -& MS & 0.106 & 0.818 & 4.750 & 0.196 & 0.874 & 0.957 & 0.979\\
 HR-Depth~\cite{hrdepth} & ResNet18& - & MS & 0.107 & 0.785 & 4.612 & 0.185 & 0.887 & 0.962 & 0.982 \\
 R-MSFM6~\cite{r-msfm} & ResNet18& - & MS & 0.111 & 0.787 & 4.625 & 0.189 & 0.882 & 0.961 & 0.981 \\
 DIFFNet~\cite{diffnet} & HRNet18& - & MS & 0.101 & 0.749 & 4.445 & 0.179 & 0.898 & 0.965 & 0.983 \\
 BRNet~\cite{brnet} & ResNet18 & -& MS & 0.099 & 0.685 & 4.453 & 0.183 & 0.885 & 0.962 & 0.983 \\
 \textbf{Ours} & ResNet18& - & MS & 0.104 & 0.717 & 4.433 & 0.178 & 0.890 & 0.964 & 0.984\\
 \textbf{Ours} & HRNet18& - & MS & \textbf{0.096} &  \textbf{0.644} & \textbf{4.166} & \textbf{0.170} & \textbf{0.905} & \textbf{0.968} & \textbf{0.985}\\
\hline 
\end{tabular}}
\begin{flushleft}
  \footnotesize\textsuperscript{} ~~~For the training type, M means monocular setting and MS means monocular plus stereo setting. In each category, the best results are in \textbf{bold}.
 \end{flushleft}
\end{center}
\vspace{-5mm}
\end{table*}

\section{Experiments}
\subsection{Datasets}
Our models are trained and evaluated on the KITTI \cite{kitti} and Cityscapes \cite{city} datasets, which are two widely used outdoor datasets for depth estimation. KITTI consists of 200 street scene videos captured with RGB cameras. For fair comparison, we use the data split in \cite{eigen2} and follow pre-processing operation in \cite{sfmlearner} to remove static frames, resulting in 39,810 monocular triplets for training and 697 for evaluation. For Cityscapes, we pre-process and train on 69,731 images from the monocular sequences following \cite{sfmlearner,manydepth}. And We evaluate on the 1,525 test images using the provided groundtruth by \cite{manydepth}.

Furthermore, we use Make3D \cite{make3d} and NYUv2 \cite{nyu} datasets for evaluation to validate the generalization ability of the models trained on KITTI. Specifically, Make3D is a small outdoor dataset containing 134 test images with aligned depth information. And NYUv2 is an indoor scene for depth estimation with 654 labeled test images. 

\subsection{Implementation Details}
For the PoseNet to estimate camera ego-motion when using monocular video sequences, we follow \cite{monodepth2} to construct it from ResNet18 \cite{resnet} and modify it to accept a pair of color images (or six channels) as input and to predict a 6-DoF relative pose. For the encoder of our DepthNet, we experiment with two kinds of backbone, ResNet18 and HRNet18 \cite{hrnet}, to validate the effectiveness of our methods. Each of the two backbones is initialized with a pretrained model on ImageNet \cite{imagenet}. Different from \cite{monodepth2}, we only predict a single depth map at the full scale to save memory and accelerate training. Our experiment framework is implemented in PyTorch \cite{pytorch}. And we train our models on KITTI with two types, monocular (M), and monocular plus stereo (MS). For each type, the models are trained for 20 epochs using AdamW \cite{adamw} optimizer, with a learning rate of $10^{-4}$ for the first 15 epochs and $10^{-5}$ for the remainder. The training resolution for KITTI is set to $640 \times 192$, the batch size is set to 10, and the maximum depth is 100m. As for Cityscapes, only the monocular (M) setting is trainable. And all the training settings are the same as KITTI, except for a different resolution of $416 \times 128$ following \cite{manydepth}. All the experiments are conducted on a single NVIDIA RTX 3090 GPU. For evaluation, we adopt error-based metrics for which lower is better (Abs Rel, Sq Rel, RMSE, RMSE log, $\rm log_{10}$) and accuracy-based metrics for which higher is better ($\delta< 1.25$ , $\delta< 1.25^2$, $\delta< 1.25^3$). And we employ median scaling before evaluation to tackle the scale ambiguity when using monocular videos.

\begin{figure*}[!t]
\centering
\includegraphics[width=6.6in]{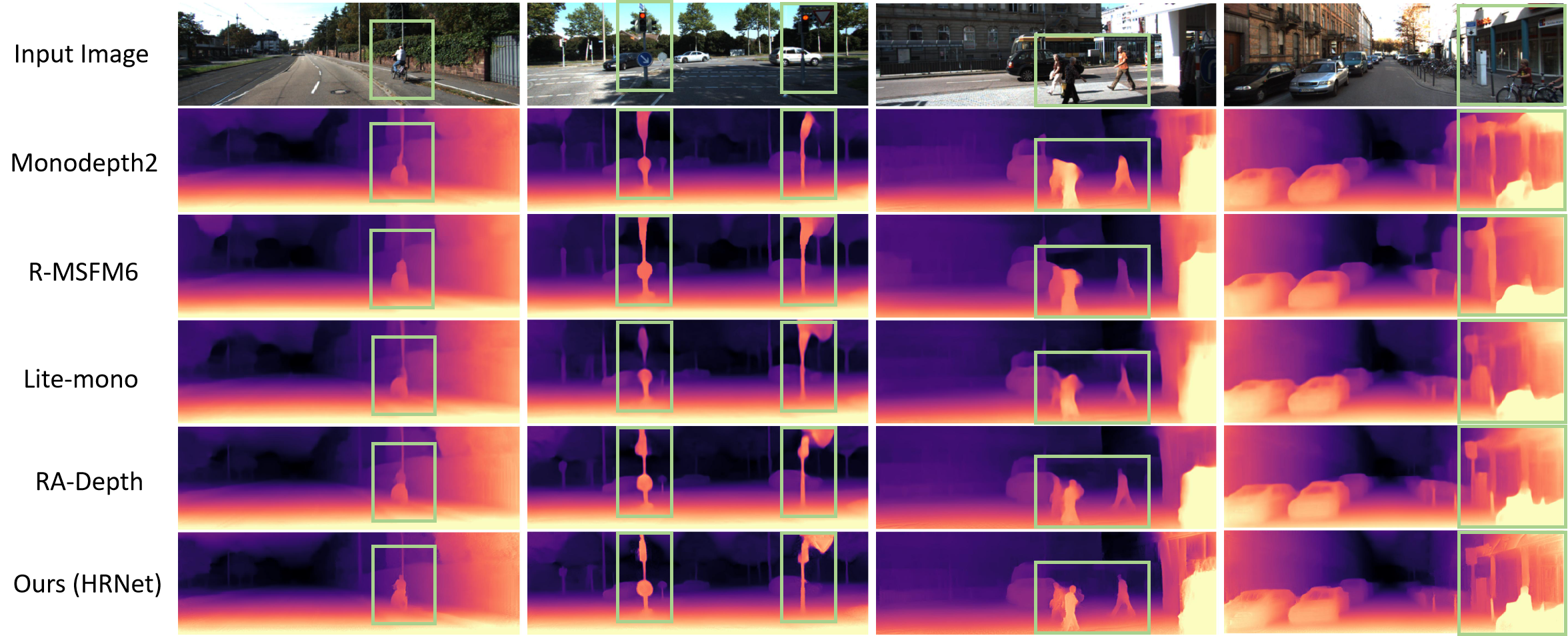}
\caption{Qualitative results on the KITTI dataset. Our model can obtain higher quality depth maps with finer depth edges compared to other methods.}
\label{vis_kitti}
\vspace{-2mm}
\end{figure*}

\subsection{Comparison on KITTI Benchmark}
For evaluation on KITTI \cite{kitti}, we cap the depth to 80m following \cite{monodepth2}. We compare the results of our models on KITTI benchmark with several recent methods, which are listed in Table~\ref{tab1}. On the raw ground truth under the monocular (M) type, our model with ResNet18 backbone already performs better than several approaches \cite{monodepth2, packnet,hrdepth,r-msfm,brnet,lite-mono,planedepth}. Furthermore, when using a more powerful backbone, HRNet18, our model can surpass them by a large margin. Besides, under the same backbone of HRNet18, our method also shows superior performance than DIFFNet \cite{diffnet} and RA-Depth \cite{radepth}. Overall, whether the training is under monocular (M) or monocular plus stereo (MS) type, our method can always achieve state-of-the-art performance. 

We also provide some qualitative comparison results on different scenes of the KITTI dataset, which are shown in Fig.~\ref{vis_kitti}. Our model is qualitatively compared with Monodepth2 \cite{monodepth2}, R-MSFM6 \cite{r-msfm}, Lite-mono \cite{lite-mono} and RA-Depth \cite{radepth}. It can be seen in the box region of each scene that our model can obtain higher quality depth maps with finer depth edges.

\begin{figure}[H]
\centering
\includegraphics[width=3.2in]{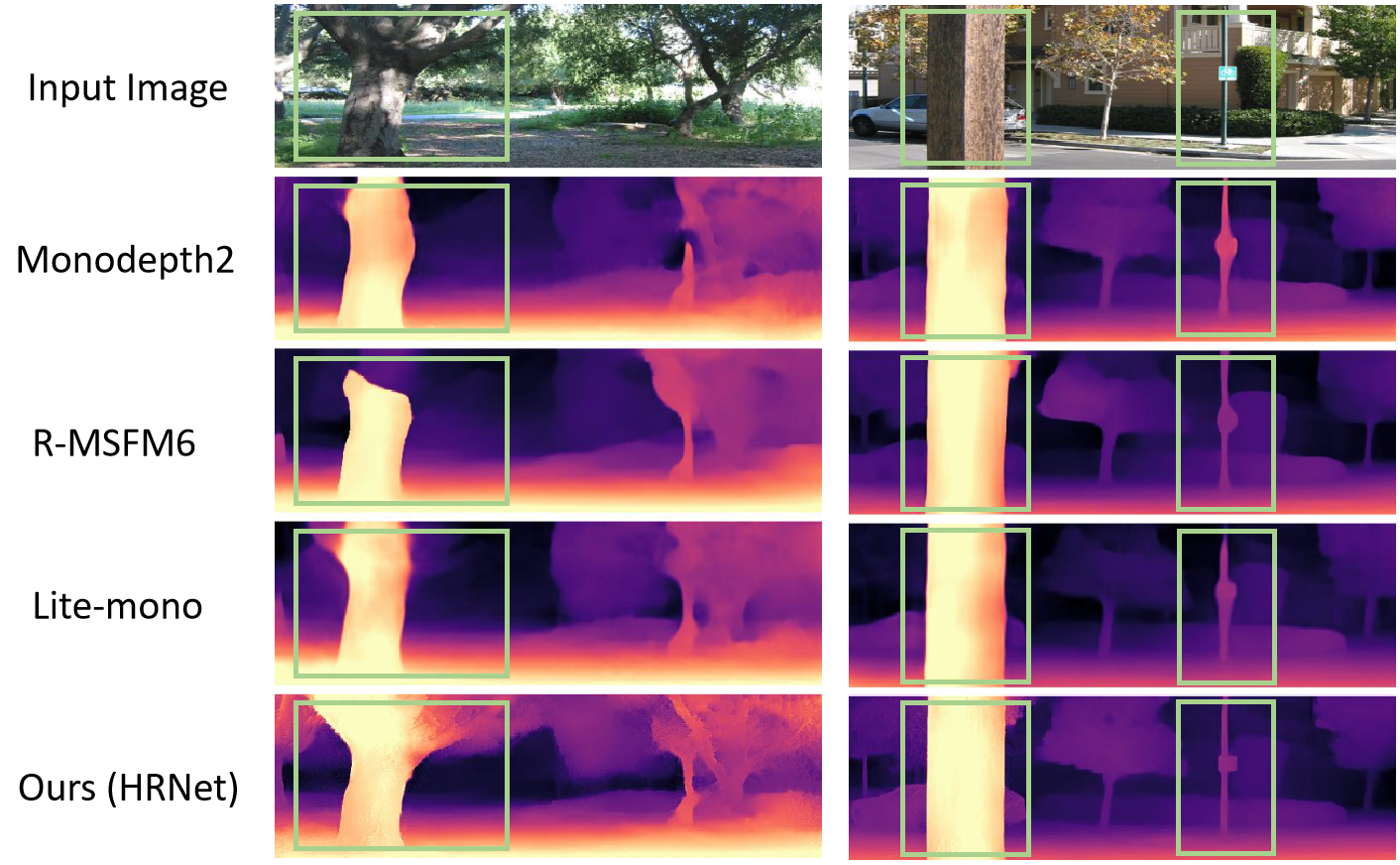}
\caption{Qualitative comparison results on the Make3D dataset.}
\label{vis_make3d}
\end{figure}

\subsection{Generalization to Make3D and NYUv2}
To validate the generalization ability of our models trained on KITTI, we directly evaluate them on Make3D \cite{make3d} and NYUv2 \cite{nyu} datasets without any fine-tuning. The maximum detph is set to 70m and 10m for Make3D and NYUv2 respectively as common practice. The results, as summarized in Table~\ref{tab2} and Table~\ref{tab3}, demonstrate that our models still outperform other methods when transfering to out-of-distribution datasets, including outdoor scenes different from KITTI and indoor scenes. It is worth noting that monocular plus stereo (MS) training has poorer generalization performance than monocular (M). We also show the qualitative results on Make3D in Fig.~\ref{vis_make3d} and It can be seen that our model can perceive objects more accurately.

\begin{table}
\begin{center}
\scriptsize
\caption{Evaluation results on Make3D DATASET}
\label{tab2}
\setlength{\tabcolsep}{1.1mm}{
\begin{tabular}{|c|c|c||c|c|c|c|}
\hline
Method & Backbone & Type & Abs Rel $\downarrow$ & Sq Rel $\downarrow$ & RMSE $\downarrow$ & $\rm log_{10}$  $\downarrow$\\
\hline
\hline
Monodepth2~\cite{monodepth2} & ResNet18 & M & 0.322 & 3.589 & 7.417 & 0.163\\
HR-Depth~\cite{hrdepth} & ResNet18 & M & 0.315 & 3.208 & 7.024 & 0.159\\
R-MSFM6~\cite{r-msfm} & ResNet18 & M & 0.334 & 3.285 & 7.212 & 0.169\\
DIFFNet~\cite{diffnet} & HRNet18 & M & 0.309 & 3.313 & 7.008 & 0.155\\
Lite-Mono~\cite{lite-mono} & Lite-Mono & M & 0.305 & 3.060 & 6.981 & 0.158 \\
BRNet~\cite{brnet} & ResNet18 & M & 0.302 & 3.133 & 7.068 & 0.156 \\
Cha et al.~\cite{Cha} & ResNet18 & M & 0.316 & 2.938 & 6.863 & 0.160 \\
\textbf{Ours} & ResNet18 & M & \textbf{0.282} & \textbf{2.626} & 6.650 & 0.149 \\
\textbf{Ours} & HRNet18 & M & 0.284 & 2.711 & \textbf{6.592} & \textbf{0.147} \\
\hline
Monodepth2~\cite{monodepth2} & ResNet18 & MS & 0.374 & 3.792 & 8.238 & 0.201\\
\textbf{Ours} & ResNet18 & MS &  0.285 & \textbf{2.659} & 6.640 &  0.148 \\
\textbf{Ours} & HRNet18 & MS & \textbf{0.282} & 2.664 & \textbf{6.546} & \textbf{0.147} \\
\hline
\end{tabular}}
\vspace{-2mm}
\end{center}
\end{table}

\begin{table}
\begin{center}
\scriptsize
\caption{Evaluation results on NYUV2 DATASET}
\label{tab3}
\setlength{\tabcolsep}{0.8mm}{
\begin{tabular}{|c|c|c||c|c|c|c|}
\hline
Method & Backbone & Type & Abs Rel $\downarrow$ & Sq Rel $\downarrow$ & RMSE $\downarrow$ & $\delta<1.25$ $\uparrow$\\
\hline
\hline
Monodepth2~\cite{monodepth2} & ResNet18 & M & 0.388 & 0.839 & 1.441 & 0.434\\
R-MSFM6~\cite{r-msfm} & ResNet18 & M & 0.372 & 0.803 & 1.344 & 0.494\\
DevNet~\cite{devnet} & ResNet50 & M & 0.333 & 0.605 & 1.142 & 0.541 \\
\textbf{Ours} & ResNet18 & M &  0.321 & 0.512 & 1.111 & 0.511  \\
\textbf{Ours} & HRNet18 & M &  \textbf{0.284} & \textbf{0.382} & \textbf{0.943} & \textbf{0.573} \\
\hline
Monodepth2~\cite{monodepth2} & ResNet18 & MS & 0.419 & 1.031 & 1.611 & 0.425\\
\textbf{Ours} & ResNet18 & MS & 0.308 & 0.448 & 1.033 &  0.508 \\
\textbf{Ours} & HRNet18 & MS & \textbf{0.301} & \textbf{0.420} & \textbf{1.010} & \textbf{0.523} \\
\hline
\end{tabular}}
\vspace{-5mm}
\end{center}
\end{table}

\subsection{Cityscapes Results}
In Table \ref{tab4} we train and evaluate our models on the Cityscapes \cite{city} dataset. We also check the generalization ability to Cityscapes of our models trained on KITTI, without any finetuning. When evaluating on Cityscapes, we cap the depth to 80m as with KITTI. Note that all models in Table \ref{tab4} are trained under monocular (M) setting. And the evaluating resolution is the same as the training ($640\times192$ for KITTI and $416\times128$ for Cityscapes). It can be seen that out models still show significant improvement in the two cases.


\begin{table}
\begin{center}
\caption{Evaluation results on CITYSCAPES DATASET}
\label{tab4}
\scriptsize
\setlength{\tabcolsep}{0.4mm}{
\begin{tabular}{|c|c|c|c||c|c|c|c|}
\hline
 Method & Backbone & Train & Test & Abs Rel $\downarrow$ & Sq Rel $\downarrow$ & RMSE $\downarrow$ & $\delta<1.25\uparrow$\\
\hline
\hline
 Monodepth2~\cite{monodepth2}& ResNet18 &  CS & CS & 0.129 & 1.569 & 6.876 & 0.849 \\
 ManyDepth~\cite{manydepth} & ResNet18 &  CS & CS & 0.114 & 1.193 & 6.223 &  \textbf{0.875}\\
 Ours & ResNet18 &  CS & CS & 0.116 & 1.107 & 6.061 &  0.868 \\
 Ours & HRNet18 &  CS & CS & \textbf{0.112} & \textbf{1.027} & \textbf{5.862} &  0.874 \\
\hline
 Monodepth2~\cite{monodepth2}& ResNet18 & K & CS & 0.164 & 1.890 & 8.985 & 0.756 \\
 Lite-Mono~\cite{lite-mono} & Lite-Mono & K & CS & 0.158 & 1.715 & 8.432 & 0.777 \\
 Ours & ResNet18 & K & CS & 0.151 & 1.633 & 8.415  & 0.778 \\
 Ours & HRNet18 & K & CS & \textbf{0.135} & \textbf{1.354} & \textbf{7.557} & \textbf{0.815} \\
\hline 
\end{tabular}}
\begin{flushleft}
  \footnotesize\textsuperscript{} ~~~CS means Cityscapes and K means KITTI.
 \end{flushleft}
\end{center}
\vspace{-3mm}
\end{table}

\subsection{Ablation Study}
In this section, we verify the effectiveness of each component in our method. All ablation experiments are trained with ResNet18 backbone and monocular setting, and are evaluated on the KITTI raw ground truth. We add each component independently to the baseline model, Monodepth2~\cite{monodepth2}, and also experiment with each component removed individually on the basis of our full method. The ablation results are listed in Table~\ref{tab5}, which demonstrate that all the components can  bring significant gains individually. One thing to note is that the self-distillation loss plays the most important part in our method. However, it cannot do without data augmentation.

\begin{table}[!t]
\begin{center}
\scriptsize
\caption{Ablation study results on KITTI raw ground truth}
\label{tab5}
\setlength{\tabcolsep}{0.8mm}{
\begin{tabular}{|c||c|c|c|c|}
\hline
Method & Abs Rel $\downarrow$ & Sq Rel $\downarrow$ & RMSE $\downarrow$ & $\delta<1.25$ $\uparrow$ \\
\hline
\hline
Baseline (Monodepth2) & 0.115 & 0.903 & 4.863 & 0.877\\
Baseline + resizing-cropping  & 0.108 & 0.778 & 4.568 & 0.885\\
Baseline + splitting-permuting  & 0.109 & 0.807 & 4.581 & 0.885 \\
Baseline + full-scale branch &  0.112 & 0.804 & 4.704 & 0.878 \\
Baseline + grid decoder &  0.111 & 0.835 & 4.752 & 0.880  \\ 
\hline 
Ours w/o resizing-cropping &  0.107 & 0.762 & 4.539 & 0.888  \\ 
Ours w/o splitting-permuting &  0.106 & 0.733 & 4.501 & 0.888  \\
Ours w/o self-distillation loss &  0.110 & 0.828 & 4.638 & 0.884  \\
Ours w/o detail-enhanced DepthNet &  0.106 & 0.730 & 4.474 & 0.887  \\
\textbf{Ours} &  \textbf{0.102} & \textbf{0.697} & \textbf{4.454} & \textbf{0.890} \\
\hline
\end{tabular}}
\begin{flushleft}
  \footnotesize\textsuperscript{} Models are trained with ResNet18 backbone and monocular (M) setting.
 \end{flushleft}
\end{center}
\end{table}

Besides, we explore more about the two data augmentation strategies and detail-enhanced DepthNet in our method.

\begin{figure}[!t]
\centering
\includegraphics[width=3.4in]{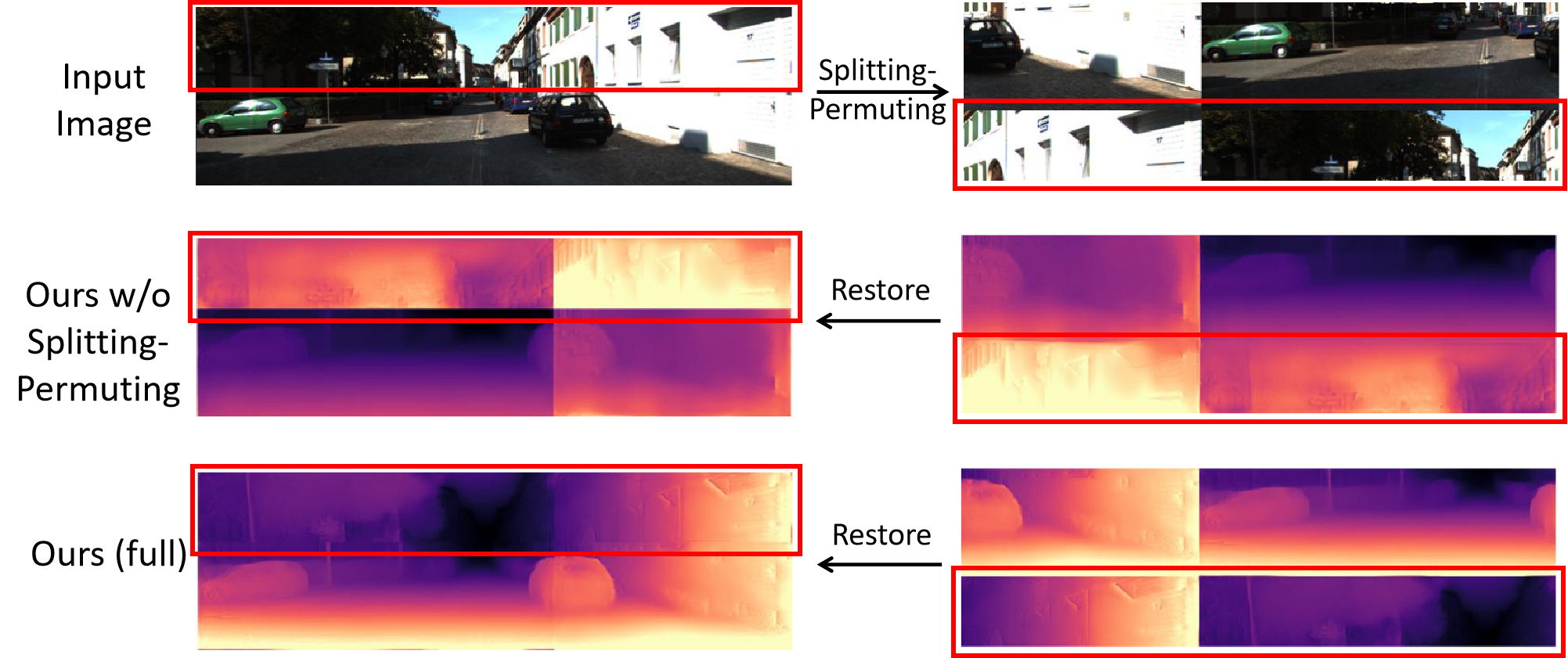}
\caption{Qualitative comparison results between our method with and without \textit{Splitting-Permuting} when processing a image augmented by this strategy. The models are trained with ResNet18 backbone and monocular setting.}
\label{ablation_sp}
\vspace{-3mm}
\end{figure}

\begin{table}[!t]
\begin{center}
\scriptsize
\caption{Resolution adaption results on KITTI raw ground truth}
\label{tab6}
\setlength{\tabcolsep}{0.3mm}{
\begin{tabular}{|c|c|c||c|c|c|c|}
\hline
Method & Backbone & Test Res. & Abs Rel $\downarrow$ & Sq Rel $\downarrow$ & RMSE $\downarrow$ & $\delta<1.25$ $\uparrow$\\
\hline
\hline
Monodepth2~\cite{monodepth2} & ResNet18 & $416\times128$ & 0.184 & 1.365 & 6.146 & 0.719\\
Ours w/o RC & ResNet18 & $416\times128$ & 0.155 & 1.109 & 6.209 & 0.780\\
Ours & ResNet18 & $416\times128$ & \textbf{0.116} & \textbf{0.803} & \textbf{5.146} & \textbf{0.855} \\
\hline
RA-Depth~\cite{radepth} & HRNet18 & $416\times128$ & 0.111 & 0.723 & 4.768 & 0.874  \\
Ours & HRNet18 & $416\times128$ & \textbf{0.107} & \textbf{0.709} & \textbf{4.721} & \textbf{0.876}   \\ 
\hline 
\hline 
Monodepth2~\cite{monodepth2} & ResNet18 & $1024\times320$ & 0.193 & 1.335 & 6.059 & 0.673\\
Ours w/o RC & ResNet18 & $1024\times320$ & 0.265 & 2.060 & 6.862 & 0.565\\
Ours & ResNet18 & $1024\times320$ & \textbf{0.108} & \textbf{0.694} & \textbf{4.272} & \textbf{0.890} \\
\hline 
RA-Depth~\cite{radepth} & HRNet18 & $1024\times320$ & \textbf{0.097} & \textbf{0.608} & 4.131 & \textbf{0.901} \\
Ours & HRNet18 & $1024\times320$ & 0.099 & 0.610 & \textbf{4.049} & \textbf{0.901}  \\ 
\hline
\end{tabular}}
\begin{flushleft}
  \footnotesize\textsuperscript{} Models are trained with $640\times192$ resolution and monocular (M) setting. RC means \textit{Resizing-Cropping} and Res. means resolution.
 \end{flushleft}
\end{center}
\vspace{-5mm}
\end{table}

\subsubsection{Resizing-Cropping} We claim that our models also have the trait of resolution adaption like RA-Depth~\cite{radepth} due to the \textit{Resizing-Cropping} augmentation. To verify this, we evaluate the models trained with $640\times192$ on two other resolutions, $416\times128$ and $1024\times320$, which is listed in Table~\ref{tab6}. When adapted to resolutions different from training, our method has comparable performance to RA-Depth. And the adaption ability becomes much worse without this augmentation.

\subsubsection{Splitting-Permuting} To better understand how \textit{Splitting-Permuting} makes sense, we estimate and then restore the depth of an image augmented by this strategy, both with and without this augmentation method, as shown in Fig.~\ref{ablation_sp}. Without \textit{Splitting-Permuting}, the model fails to perceive the scene structure when the image context is broken, especially in the regions (see the box area in Fig.~\ref{ablation_sp}) where there are no roads to find clues from. When incorporating \textit{Splitting-Permuting}, our method can restore depths of these regions even with context broken and clues missing. That accounts for our models' generalization capacity to diverse scenes.

\subsubsection{Detail-enhanced DepthNet} 
\begin{figure}[H]
\centering
\includegraphics[width=3.2in]{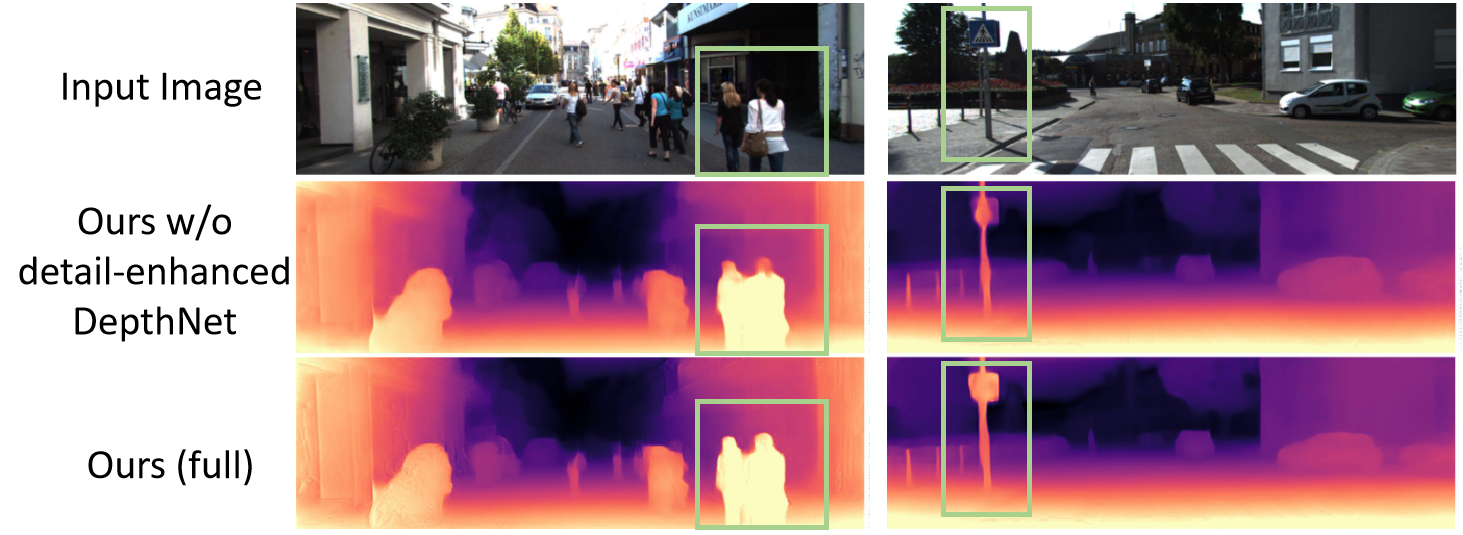}
\caption{Qualitative comparison results between our full method and that without detail-enhanced DepthNet. Models are trained with ResNet18 backbone and monocular setting.}
\label{ablation_arch}
\end{figure}

We additionally provide the qualitative comparison results between our full method and that without detail-enhanced DepthNet, as shown in Fig.~\ref{ablation_arch}. It is obviously that detail-enhanced DepthNet can help to restore more fine details in the estimated depth maps.

\section{Conclusion}
In conclusion, this paper presents a novel approach for self-supervised monocular depth estimation which utilizes data augmentation and self-distillation techniques. By incorporating \textit{Resizing-Cropping} and \textit{Splitting-Permuting}, we can fully and effectively exploit the potential of training datasets and improve the generalization ability of the models. Moreover, we introduce the detail-enhanced DepthNet with an additional full-scale branch in the encoder and a grid decoder, which significantly improves the restoration of fine details in depth maps.  Experimental results demonstrate the effectiveness and superior generalization capability of our method.


\ifCLASSOPTIONcaptionsoff
  \newpage
\fi


\begin{thebibliography}{1}
\bibliographystyle{IEEEtran}

\bibitem{eigen}
 D. Eigen, C. Puhrsch, and R. Fergus, ``Depth map prediction from a single image using a multi-scale deep network,'' in \textit{Proc. Adv. Neural Inf. Process. Syst.}, 2014, pp. 2366--2374.

\bibitem{liu}
 F. Liu, C. Shen, and G. Lin, ``Deep convolutional neural fields for depth estimation from a single image,'' in \textit{Proc. IEEE Conf. Comput. Vis. Pattern Recognit.}, 2015, pp. 5162--5170.

\bibitem{li}
 B. Li, C. Shen, Y. Dai, A. Hengel, and M. He, ``Depth and surface normal estimation from monocular images using regression on deep features and hierarchical CRFs,'' in \textit{Proc. IEEE Conf. Comput. Vis. Pattern Recognit.}, 2015, pp. 1119--1127.

\bibitem{laina}
 I. Laina, C. Rupprecht, V. Belagiannis, F. Tombari, and N. Navab, ``Deeper depth prediction with fully convolutional residual networks,'' in \textit{Proc. IEEE Int. Conf. 3D Vis.}, 2016, pp. 239--248.

\bibitem{fu}
 H. Fu, M. Gong, C. Wang, K. Batmanghelich, and D. Tao, ``Deep ordinal regression network for monocular depth estimation,'' in \textit{Proc. IEEE Conf. Comput. Vis. Pattern Recognit.}, 2018, pp. 2002--2011.

\bibitem{eigen2}
 D. Eigen and R. Fergus, ``Predicting depth, surface normals and semantic labels with a common multi-scale convolutional architecture,'' in \textit{Proc. IEEE Int. Conf. Comput. Vis.}, 2015, pp. 2650--2658.

\bibitem{etc}
 H. Zhang, C. Shen, Y. Li, Y. Cao, Y. Liu, and Y. Yan, ``Exploiting temporal consistency for real-time video depth estimation,'' in \textit{Proc. IEEE Int. Conf. Comput. Vis.}, 2019, pp. 1725--1734.
 
\bibitem{dsl}
 J. Liu, L. Kong, and Y. Jie, ``Designing and searching for lightweight monocular depth network,'' in \textit{Proc. Int. Conf. Neural Inf. Process.}, 2021, pp. 477--488.
 
\bibitem{Garg}
 R. Garg, V. K. B.G., G. Carneiro, and I. Reid, ``Unsupervised CNN for single view depth estimation: Geometry to the rescue,'' in \textit{Proc. Eur. Conf. Comput. Vis.}, 2016, pp. 740--756.

\bibitem{Godard}
 C. Godard, O. M. Aodha, and G. J. Brostow, ``Unsupervised monocular depth estimation with left-right consistency,'' in \textit{Proc. IEEE Conf. Comput. Vis. Pattern Recognit.}, 2017, pp. 6602--6611.
 
\bibitem{sfmlearner}
 T. Zhou, M. Brown, N. Snavely, and D. G. Lowe, ``Unsupervised learning of depth and ego-motion from video,'' in \textit{Proc. IEEE Conf. Comput. Vis. Pattern Recognit.}, 2017, pp. 6612--6619.

\bibitem{MDFlow}
L. Kong and Y. Jie, ``MDFlow: Unsupervised optical flow learning by reliable mutual knowledge distillation,'' in \textit {IEEE Trans. Circuits Syst. Video Tech.}, 2023, pp. 677--688.
  
\bibitem{geonet}
 Z. Yin and J. Shi, ``GeoNet: Unsupervised learning of dense depth, optical flow and camera pose,'' in \textit{Proc. IEEE Conf. Comput. Vis. Pattern Recognit.}, 2018, pp. 1983--1992.

\bibitem{sdo}
 M. Klingner, J. Termöhlen, J. Mikolajczyk, and T. Fingscheidt, ``Self-supervised monocular depth estimation: Solving the dynamic object problem by semantic guidance,'' in \textit{Proc. Eur. Conf. Comput. Vis.}, 2020, pp. 582--600.

\bibitem{poggi}
 M. Poggi, F. Aleotti, F. Tosi, and S. Mattoccia, ``On the uncertainty of self-supervised monocular depth estimation,'' in \textit{Proc. IEEE Conf. Comput. Vis. Pattern Recognit.}, 2020, pp. 3224--3234. 

\bibitem{monodepth2}
 C. Godard, O. Mac Aodha, M. Firman, and G. J. Brostow, ``Digging into self-supervised monocular depth estimation,'' in \textit{Proc. IEEE Int. Conf. Comput. Vis.}, 2019, pp. 3828--3838.
 
\bibitem{packnet}
 V. Guizilini, R. Ambrus, S. Pillai, A. Raventos, and A. Gaidon, ``3D packing for self-supervised monocular depth estimation,'' in \textit{Proc. IEEE Conf. Comput. Vis. Pattern Recognit.}, 2020, pp. 2482--2491.
 
\bibitem{hrdepth}
 X. Lyu  et al., ``HR-Depth: High resolution self-supervised monocular depth estimation,'' in \textit{Proc. AAAI Conf. Artif. Intell.}, 2021, pp. 2294--2301.

\bibitem{r-msfm}
 Z. Zhou, X. Fan, P. Shi, and Y. Xin, ``R-MSFM: Recurrent multi-scale feature modulation for monocular depth estimating,'' in \textit{Proc. IEEE Int. Conf. Comput. Vis.}, 2021, pp. 12757--12766.
 
\bibitem{adaadepth}
 V. Kaushik, K. Jindgar, and B. Lall, ``ADAADepth: Adapting data augmentation and attention for self-supervised monocular depth estimation,'' \textit{IEEE Robot. Automat. Lett.}, vol. 6, no. 4, pp. 7791--7798, Oct. 2021.
 
\bibitem{brnet}
 W. Han, J. Yin, X. Jin, X. Dai, and J. Shen, ``BRNet: Exploring comprehensive features for monocular depth estimation,'' in \textit{Proc. Eur. Conf. Comput. Vis.}, 2022, pp. 586--602.

\bibitem{Cha}
  G. Cha, H.-D. Jang, and D. Wee, ``Self-supervised depth estimation with isometric-self-sample-based learning,'' \textit{IEEE Robot. Automat. Lett.}, vol. 8, no. 4, pp. 2173--2180, Apr. 2023.

\bibitem{lite-mono}
 N. Zhang, F. Nex, G. Vosselman, and N. Kerle, ``Lite-Mono: A lightweight CNN and Transformer architecture for self-supervised monocular depth estimation,'' in \textit{Proc. IEEE Conf. Comput. Vis. Pattern Recognit.}, 2023, pp. 18537--18546.

\bibitem{planedepth}
 R. Wang, Z. Yu, and S. Gao, ``PlaneDepth: Self-supervised depth estimation via orthogonal planes,'' in \textit{Proc. IEEE Conf. Comput. Vis. Pattern Recognit.}, 2023, pp. 21425--21434.

\bibitem{devnet}
 K. Zhou et al., ``DevNet: Self-supervised monocular depth learning via density volume construction,'' in \textit{Proc. Eur. Conf. Comput. Vis.}, 2022, pp. 125--142.
 
\bibitem{diffnet}
 H. Zhou, D. Greenwood, and S. Taylor, ``Self-supervised monocular depth
estimation with internal feature fusion,'' in \textit{Proc. Bri. Mach. Vis. Conf.}, 2021.
 
\bibitem{radepth}
 M. He, L. Hui, Y. Bian, J. Ren, J. Xie, and J. Yang, ``RA-Depth: Resolution adaptive self-supervised monocular depth estimation,'' in \textit{Proc. Eur. Conf. Comput. Vis.}, 2022, pp. 565--581.
 
\bibitem{scdepth}
 J. Bian et al., ``Unsupervised scale-consistent depth and ego-motion learning from monocular video,'' in \textit{Proc. Adv. Neural Inf. Process. Syst.}, 2019, pp. 35--45.

\bibitem{epcdepth}
 R. Peng, R. Wang, Y. Lai, L. Tang, and Y. Cai, ``Excavating the potential capacity of self-supervised monocular depth estimation,'' in \textit{Proc. IEEE Int. Conf. Comput. Vis.}, 2021, pp. 15540--15549.
 
\bibitem{Pilzer}
 A. Pilzer, S. Lathuiliere, N. Sebe, and E. Ricci, ``Refine and distill: Exploiting cycle-inconsistency and knowledge distillation for unsupervised monocular depth estimation,'' in \textit{Proc. IEEE Conf. Comput. Vis. Pattern Recognit.}, 2019, pp. 9760--9769.

\bibitem{Bello}
 J. Bello and M. Kim, ``Self-supervised deep monocular depth estimation with ambiguity boosting,'' in \textit{IEEE Trans. Pattern Anal. Mach. Intell.}, vol. 44, no. 12, pp. 9131--9149, Dec. 2022.
 
\bibitem{direct}
 C. Wang, J. M. Buenaposada, R. Zhu, and S. Lucey, ``Learning depth from monocular videos using direct methods,'' in \textit{Proc. IEEE Conf. Comput. Vis. Pattern Recognit.}, 2018, pp. 2022--2030.

\bibitem{superdepth}
 S. Pillai, R. Ambru¸s, and A. Gaidon, ``Superdepth: Self-supervised, super-resolved monocular depth estimation,'' in \textit{Proc. IEEE Int. Conf. Robot. Automat}, 2019, pp. 9250--9256.
 
\bibitem{jiang2020}
 H. Jiang, L. Ding, Z. Sun, and R. Huang, ``Dipe: Deeper into photometric errors for unsupervised learning of depth and ego-motion from monocular videos,'' in \textit{Proc. IEEE. Int. Conf. Intell. Robot. Syst.}, 2020, pp. 10061--10067.
 
\bibitem{li2021}
 H. Li, A. Gordon, H. Zhao, V. Casser, and A. Angelova, ``Unsupervised monocular depth learning in dynamic scenes,'' in \textit{n Proc. Conf. Robot Learn.}, 2021, pp. 1908--1917.

\bibitem{Wagstaff}
 B. Wagstaff, V. Peretroukhin, and J. Kelly, ``On the coupling of depth and egomotion networks for self-supervised structure from motion,'' \textit{IEEE Robot. Automat. Lett.}, vol. 7, no. 3, pp. 6766--6773, Jul. 2022.

\bibitem{manydepth}
 J. Watson, O. Mac Aodha, V. Prisacariu, G. Brostow, and M. Firman, ``The temporal opportunist: Self-supervised multi-frame monocular depth,'' \textit{Proc. IEEE Conf. Comput. Vis. Pattern Recognit.}, 2021, pp. 1164--1174.
 
\bibitem{unet}
 O. Ronneberger, P. Fischer, and T. Brox, ``U-Net: Convolutional networks for biomedical image segmentation,'' in \textit{Proc. Med. Image Comput. and Computer-Assisted Interv.}, 2015, pp. 234--241.

\bibitem{kitti}
 A. Geiger, P. Lenz, C. Stiller, and R. Urtasun, ``Vision meets robotics: The kitti dataset,'' \textit{Int. J. Robot. Res.}, vol. 32, no. 11, pp. 1231--1237, Sept. 2013.

\bibitem{city}
 M. Cordts et al., ``The cityscapes dataset for semantic urban scene understanding,'' \textit{Proc. IEEE Conf. Comput. Vis. Pattern Recognit.}, 2016, pp. 3213--3223.

\bibitem{make3d}
 A. Saxena, M. Sun, and A. Y. Ng, ``Make3d: Learning 3d scene structure from a single still image,'' in \textit{IEEE Trans. Pattern Anal. Mach. Intell.}, vol. 31, no. 5, pp. 824--840, May. 2009.

\bibitem{nyu}
 N. Silberman et al., ``Indoor segmentation and support inference from RGBD images,'' in \textit{Proc. Eur. Conf. Comput. Vis.}, 2012, pp. 746--760.
 
\bibitem{ssim}
 Z. Wang, A. C. Bovik, H. R. Sheikh, and E. P. Simoncelli, ``Image quality assessment: From error visibility to structural similarity,'' \textit{IEEE Trans. Image Process.}, vol. 13, no. 4, pp. 600--612, Apr. 2004.

\bibitem{gridnet}
 D. Fourure, R. Emonet, E. Fromont, D. Muselet, A. Tremeau, and C. Wolf, ``Residual conv-deconv grid network for semantic segmentation,'' in \textit{Proc. Bri. Mach. Vis. Conf.}, 2017.
 
\bibitem{resnet}
 K. He, X. Zhang, S. Ren, and J. Sun, ``Deep residual learning for image recognition,'' in \textit{Proc. IEEE Conf. Comput. Vis. Pattern Recognit.}, 2016, pp. 770--778.

\bibitem{hrnet}
 J. Wang et al., ``Deep high-resolution representation learning for visual recognition,'' \textit{IEEE Trans. Pattern Anal. Mach. Intell.}, vol. 43, no. 10, pp. 3349--3364, Oct. 2021.
 
\bibitem{imagenet}
 J. Deng et al., ``Imagenet: A large-scale hierarchical image database,'' in \textit{Proc. IEEE Conf. Comput. Vis. Pattern Recognit.}, 2009, pp. 248--255.

\bibitem{pytorch}
 A. Paszke et al., ``PyTorch: An imperative style, high-performance deep learning library,'' in \textit{Proc. Adv. Neural Inf. Process. Syst.}, 2019, pp. 8026--8037.

\bibitem{adamw}
 I. Loshchilov and F. Hutter, ``Decoupled weight decay regularization,'' in \textit{Proc. Int. Conf. Learn. Representations}, 2019.

\end{thebibliography}
\end{document}